\documentclass[10pt,twocolumn,letterpaper]{article}

\usepackage[pagenumbers]{cvpr}
\usepackage{graphicx}
\usepackage{amsmath}
\usepackage{amssymb}
\usepackage{booktabs}
\usepackage{pifont}
\usepackage{xcolor}
\definecolor{cvprblue}{rgb}{0.21,0.49,0.74}
\usepackage[breaklinks,colorlinks,allcolors=cvprblue]{hyperref}
\def\confName{arXiv}
\def\confYear{2026}

\newcommand{\cmark}{\ding{51}}
\title{Aligned Consensus Teaching for Label-Efficient Oriented Object Detection in Weakly-Aligned Visible-Infrared Imagery}
\author{
    {\normalsize Qi Ming$^{1}$ \quad Xiaxin Yuan$^{2}$ \quad Jiahuan Zhou$^{3}$ \quad
    Jiangmeng Li$^{4}$ \quad Xudong Zhao$^{5}$}\\
    {\normalsize Zhanchao Huang$^{6}$ \quad Juan Fang$^{1}$ \quad
    Shaoguang Huang$^{7}$ \quad Aleksandra Pizurica$^{8}$}\\[3pt]
    {\small $^{1}$Beijing University of Technology \quad
    $^{2}$University of Science and Technology of China \quad
    $^{3}$Peking University}\\
    {\small $^{4}$Institute of Software, Chinese Academy of Sciences \quad
    $^{5}$Beijing Institute of Technology}\\
    {\small $^{6}$Fuzhou University \quad
    $^{7}$China University of Geosciences Wuhan}\\
    {\small $^{8}$Telecommunications and Information Processing, Ghent University}\\
    {\tt\small chaser.ming@gmail.com \quad xiaxinyuan@mail.ustc.edu.cn}\\
    {\tt\small Aleksandra.Pizurica@UGent.be}
}

\begin{document}

\maketitle

\begin{abstract}
Visible-infrared object detection (VIOD) detects objects with oriented bounding boxes from paired visible and infrared images.
Existing methods depend on costly dual-modality annotations.
Semi-supervised learning can reduce this burden, but extending it from single-modal detection to VIOD is challenging.
In the practical image-pair-level setting considered here, only a few pairs are labeled in both modalities, while the rest are completely unlabeled.
This limited supervision creates three challenges: (i) too few labeled boxes for robust cross-modal alignment; (ii) pseudo-label errors caused by branch-wise misses accumulate during self-training; and (iii) tail-class annotations become critically scarce as the labeling budget decreases.
We propose Aligned Consensus Teacher (ACT) for label-efficient VIOD in this setting.
Its Cycle-Consistent Region Alignment (CRA) combines cycle consistency and sparse anchors with reliability-weighted regional matching.
Cross-Modal Consensus Mean-Teacher (CMC-MT) forms consensus pseudo labels under pair-preserving views to recover branch-wise misses and supervise unlabeled pairs.
Text-Guided Cross-Modal Instance Augmentation (TG-CMIA) uses a vision-language scene prior to compose tail-class instance pairs while preserving RGB--IR offsets.
To the best of our knowledge, ACT is the first framework to study semi-supervised VIOD under this image-pair-level setting.
Experiments on DroneVehicle and VEDAI show consistent gains across annotation ratios.
With 10\% labeled pairs on DroneVehicle, ACT reaches 94.3\% of the mAP obtained by the same detector under full supervision.
Code and models will be available on GitHub to facilitate future work.
\end{abstract}

\section{Introduction}

Object detection in aerial images supports a wide range of applications, including traffic surveillance, urban management, and disaster rescue~\cite{sun2022dronevehicle,xia2018dota}.
Aerial objects appear at arbitrary orientations, so detectors predict rotated bounding boxes~\cite{ding2019roitransformer,han2021s2anet,xie2021orientedrcnn}.
However, detectors that use only visible (RGB) images often degrade at night or under adverse weather~\cite{yuan2022tsfadet}.
Infrared (IR) images capture the radiated heat of objects and reveal clear silhouettes even in low light, so the two modalities are complementary~\cite{hwang2015multispectral,yuan2024c2former,sun2022dronevehicle}.
By jointly exploiting them, visible-infrared object detection (VIOD) works under both daytime and low-light conditions.

\begin{figure}[!t]\centering
\includegraphics[width=\columnwidth]{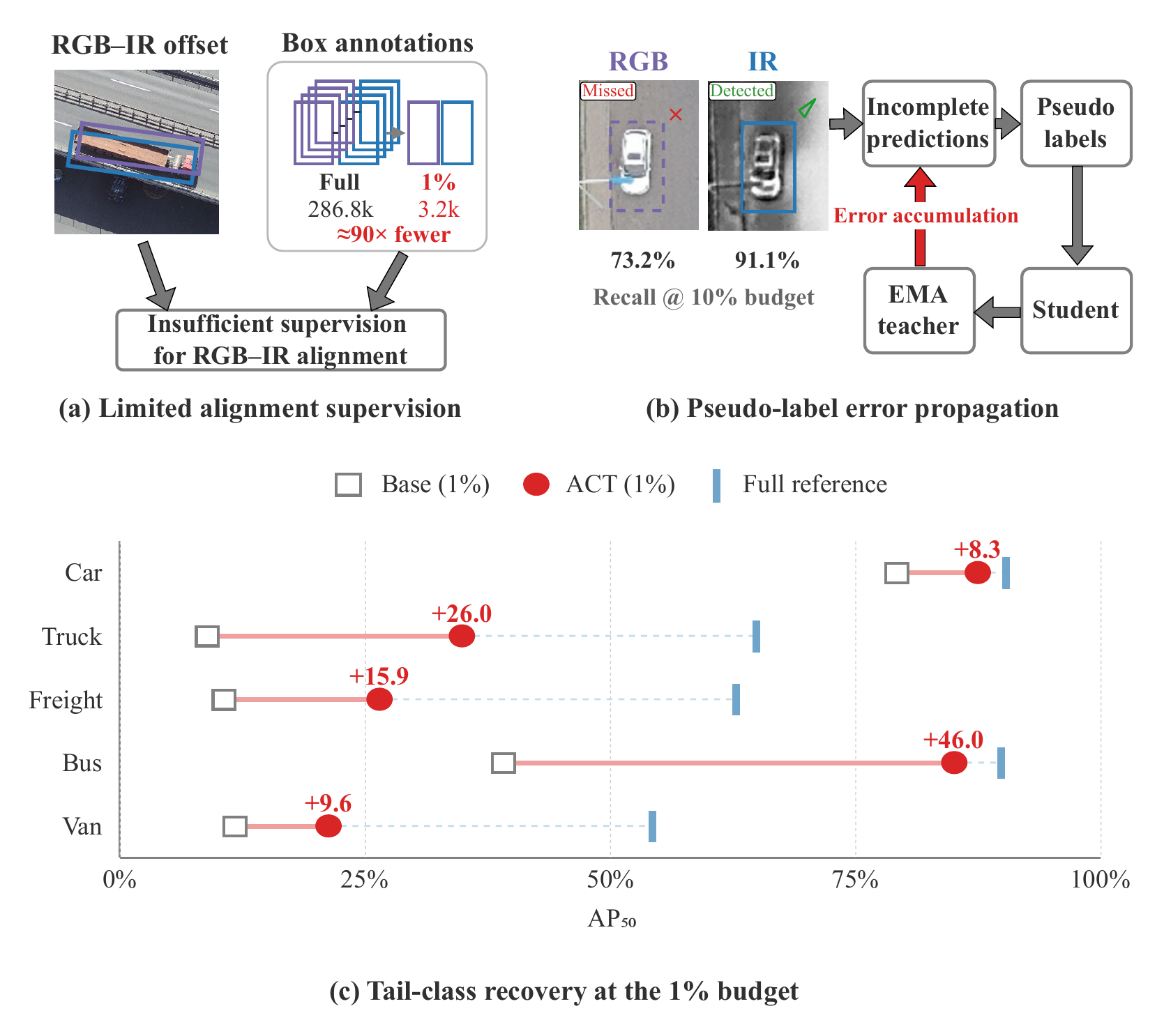}
\caption{Motivation on DroneVehicle. (a) A $1\%$ labeling budget provides about $90\times$ fewer matched boxes for RGB--IR alignment. (b) A branch-wise miss yields incomplete pseudo labels whose errors propagate through EMA self-training. (c) ACT recovers class-wise AP toward the full-supervision reference at the $1\%$ budget.}
\label{fig:teaser}
\end{figure}

Existing VIOD detectors are mostly fully supervised and thus rely on abundant annotations for both modalities~\cite{yuan2024c2former,yuan2022tsfadet,sun2022dronevehicle}.
Such annotations are costly to obtain, because the visible and infrared images must be annotated separately, which roughly doubles the labeling effort.
Learning VIOD detectors from only a few labeled pairs is therefore of great practical value.
Semi-supervised learning offers a natural way to reduce this cost, because it trains the detector with a few image pairs labeled in both modalities and abundant completely unlabeled pairs~\cite{xu2021softteacher,hua2023sood}.

However, extending semi-supervised learning from single-modal object detection to VIOD remains under-explored.
Under this limited paired supervision, three challenges arise.
(i) \textbf{Insufficient Cross-Modal Alignment Supervision}: The two sensors differ in field of view and imaging time, so the same object shows intrinsic position and size deviations~\cite{yuan2022tsfadet,yuan2024c2former}. Only labeled pairs provide matched boxes for regional correspondence. On DroneVehicle, reducing the labeling budget to $1\%$ decreases their number from 286.8k to 3.2k, leaving insufficient supervision for robust cross-modal alignment (Figure~\ref{fig:teaser}(a)).
(ii) \textbf{Accumulated Pseudo-Label Noise}: Visibility differences between RGB and IR cause branch-wise misses and unequal recall across the two modalities. In Figure~\ref{fig:teaser}(b), RGB recall is 73.2\%, whereas IR recall reaches 91.1\% at the $10\%$ budget. Such misses produce incomplete pseudo labels whose errors can accumulate through EMA self-training.
(iii) \textbf{Tail-Class Annotation Scarcity}: Aerial vehicles follow a long-tailed distribution. As the labeled subset shrinks, the absolute number of annotated tail instances drops sharply even when the class proportions remain similar. Figure~\ref{fig:teaser}(c) shows the resulting performance gap. When the same supervised detector is trained with $1\%$ rather than all labels, AP drops substantially more on the minority categories than on car. Tail classes consequently receive too little supervision to learn their diverse appearances and orientations~\cite{yu2021dshnet,zhang2022acrst,li2023cissod,tan2021eqlv2}.

To address these challenges, we propose Aligned Consensus Teacher (ACT), a label-efficient framework for image-pair-level semi-supervised VIOD.
First, a Cycle-Consistent Region Alignment module combines target-free cycle consistency with sparse cross-modal anchors and reliability-weighted matching to regularize regional correspondence under limited paired supervision.
Next, a Cross-Modal Consensus Mean-Teacher forms consensus pseudo labels from the CRA-regularized fused branch under pair-preserving views to recover branch-wise misses and supervise unlabeled pairs.
Finally, a Text-Guided Cross-Modal Instance Augmentation composes minority-class instance pairs in both modalities to enrich tail-class supervision.
Extensive experiments on DroneVehicle and VEDAI show that ACT consistently outperforms state-of-the-art methods under all annotation budgets. With only 10\% labeled pairs on DroneVehicle, it reaches 94.3\% of the mAP obtained by the same detector under full supervision.
The main contributions are summarized as follows:
\begin{itemize}
\item We study image-pair-level semi-supervised VIOD and propose ACT to learn from a few jointly labeled RGB--IR pairs and many completely unlabeled pairs. To the best of our knowledge, ACT is the first framework for this setting.
\item To regularize cross-modal correspondence under limited paired supervision, we design a Cycle-Consistent Region Alignment (CRA) module that combines target-free cycle consistency, sparse cross-modal anchors, and reliability-weighted regional matching.
\item To reduce pseudo-label errors from branch-wise misses, we develop a Cross-Modal Consensus Mean-Teacher (CMC-MT) that derives reliable pseudo labels from cross-modal consensus under pair-preserving views.
\item To address tail-class annotation scarcity, we introduce a Text-Guided Cross-Modal Instance Augmentation (TG-CMIA) strategy that composes minority-class instance pairs coherently across both modalities under a vision-language scene prior.
\end{itemize}

\section{Related Work}

\textbf{Visible-Infrared Object Detection.}
Oriented object detection is essential in aerial scenes, where objects appear at arbitrary angles~\cite{sun2022dronevehicle}.
Existing methods adapt two-stage proposal pipelines to rotated regions~\cite{ren2015fasterrcnn,ding2019roitransformer,xie2021orientedrcnn} or use one-stage dense detectors~\cite{lin2017focal,han2021s2anet}, with later advances in rotation-equivariant representation and geometry-aware localization~\cite{han2021redet,ming2026herodet,yang2021r3det,yang2021gwd}.
These RGB-only detectors can degrade under poor illumination.
Visible-infrared detectors instead exploit complementary sensing through cross-modal alignment~\cite{zhang2019arcnn,yuan2022tsfadet}, feature calibration~\cite{yuan2024c2former}, or state-space fusion~\cite{zhou2025dmm,dong2025fusionmamba}.
Because the paired views are weakly aligned, they learn spatial correspondence from densely labeled boxes.
Most therefore remain fully supervised and cannot obtain such correspondence supervision when only a few image pairs are labeled.

\textbf{Semi-Supervised Object Detection.}
Semi-supervised object detection trains on a few labeled images and many unlabeled images~\cite{xu2021softteacher}.
Most methods follow mean-teacher self-training, where an EMA teacher generates pseudo labels for a student~\cite{liu2021unbiased,xu2021softteacher,zhou2022denseteacher}; this paradigm has also been extended to oriented objects~\cite{hua2023sood,wang2025mcl}.
Subsequent methods improve pseudo-label quality through detection consistency, dense supervision, localization-aware scoring, adaptive assignment, or label polishing~\cite{tang2021humble,li2022pseco,zhou2022denseteacher,li2022rethinkingpseudo,wang2023consistent,zhang2023polishing}.
However, they assume a single modality and do not model cross-modal displacement or complementary branch-wise misses.
Sparsely annotated multimodal detection instead studies missing instances within an image~\cite{lee2025sampd} or annotations available in only one modality~\cite{jin2025dodsa}.
These settings differ from image-pair-level semi-supervised VIOD, where a few pairs are labeled in both modalities, the remaining pairs are completely unlabeled, and augmentation must preserve RGB--IR pairing.

\section{Methodology}

\subsection{Overview}
\begin{figure*}[t]
\centering
\includegraphics[width=\textwidth]{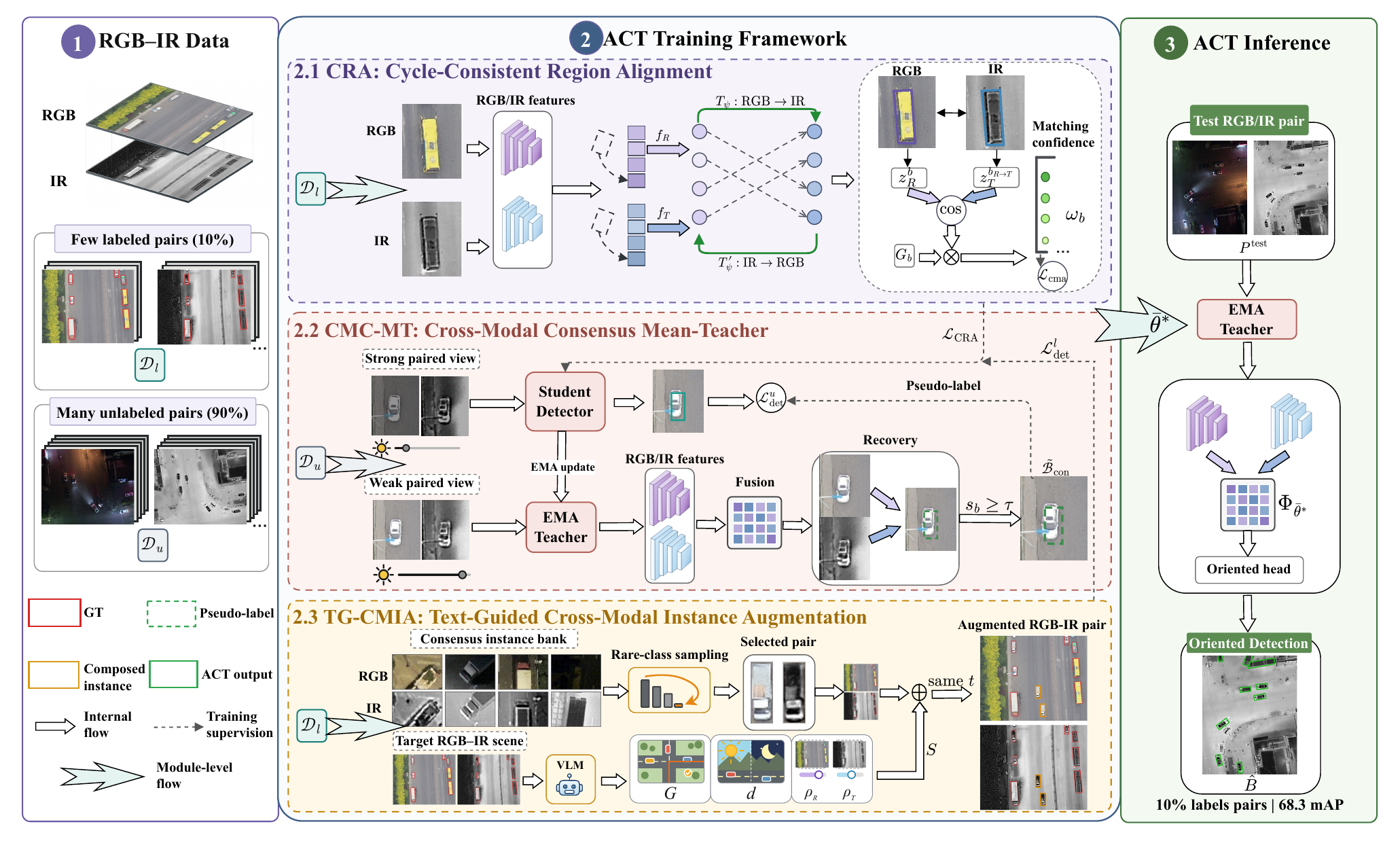}
\caption{Training and inference overview of ACT. CRA learns regional correspondence from labeled pairs, CMC-MT converts unlabeled pairs into consensus supervision, and TG-CMIA composes tail-class RGB--IR instance pairs on the labeled branch. Only the fused EMA teacher is retained for inference.}
\label{fig:framework}
\end{figure*}

Figure~\ref{fig:framework} illustrates ACT for image-pair-level semi-supervised VIOD.
CRA regularizes regional correspondence on labeled RGB--IR pairs, CMC-MT generates consensus pseudo labels for unlabeled pairs under pair-preserving views, and TG-CMIA composes tail-class pairs on the labeled branch.
The student learns from these supervised and unsupervised objectives.
At inference, only the fused EMA teacher and oriented detection head are retained.

\subsection{Cycle-Consistent Region Alignment}
RGB and infrared images exhibit object-level displacement, while limited labeled boxes are insufficient to learn robust cross-modal alignment.
CRA addresses this problem by coupling cycle consistency with sparse correspondence anchors and reliability-weighted regional matching.

\noindent\textbf{Cycle-Consistent Alignment with Sparse Anchors.}
Let $f_R$ and $f_T$ denote the RGB and infrared feature maps, let $b=(c_x,c_y,w,h,\theta)$ be an RGB region, and define $r_m(b)=\mathrm{RoI}(f_m,b)$ using RoI-aligned regional features~\cite{he2017maskrcnn}.
A lightweight forward head $T_{\psi}$ predicts its center and scale offset from the regional features of both modalities.
A reverse head $T_{\psi'}$ then maps the transformed region back to the RGB frame:
\begin{equation}
\begin{split}
\Delta_b^{R\rightarrow T}
&=T_{\psi}\!\left(
  [r_R(b),r_T(b)]
  \right),\\
b_{R\rightarrow T}
&=\mathcal{T}(b,\Delta_b^{R\rightarrow T}),\\
b_{R\rightarrow T\rightarrow R}
&=\mathcal{T}\!\left(
  b_{R\rightarrow T},
  T_{\psi'}\!\left(
  [r_T(b_{R\rightarrow T}),
  r_R(b_{R\rightarrow T})]\right)
  \right).
\end{split}
\raisetag{2\baselineskip}
\label{eq:offset}
\end{equation}
Here, $\mathcal{T}$ adjusts the center and scale while retaining the orientation.
The round trip imposes cycle consistency without a matched infrared box for every RGB source region.
\mbox{Class-consistent} RGB--IR matches define $\mathcal{A}$.
\begin{subequations}\label{eq:cra_constraints}
\begin{align}
\mathcal{L}_{\mathrm{cyc}}
&=\frac{1}{|\mathcal{R}|}
  \sum_{b\in\mathcal{R}}
  \left\|
  (b_{R\rightarrow T\rightarrow R}-b)_{1:4}
  \right\|_1,
\label{eq:cyc}\\
\mathcal{L}_{\mathrm{align}}
&=\frac{1}{\max(|\mathcal{A}|,1)}
  \sum_{b\in\mathcal{A}}
  \ell_{\mathrm{SL1}}
  \!\left(\Delta_b^{R\rightarrow T}-\Delta_b^{*}\right).
\label{eq:align}
\end{align}
\end{subequations}
Here, $\mathcal{R}$ contains RGB ground-truth regions from the labeled pairs, and $\Delta_b^{*}$ is encoded from a matched infrared box.
Equation~(\ref{eq:cyc}) is target-free because it needs no matched infrared offset for every source region; this does not imply optimization on the unlabeled branch.
Since a cycle alone admits mutually consistent yet incorrect mappings, the sparse-anchor loss in Eq.~(\ref{eq:align}) resolves this ambiguity and prevents transformation drift.

\noindent\textbf{Reliability-Weighted Regional Matching.}
Even with geometric constraints, appearance agreement varies across regions because illumination and sensing conditions affect the two modalities differently.
We denote the regional features before and after mapping by
$z_R^b=\mathrm{RoI}(f_R,b)$ and
$z_T^{b_{R\rightarrow T}}=\mathrm{RoI}(f_T,b_{R\rightarrow T})$.
A coarse occupancy score $G_b\in[0,1]$ from the offline scene prior refines their visual agreement and yields the regional confidence
\begin{equation}
\begin{array}{rl}
\omega_b^{\mathrm{vis}}
&=\sigma\!\left(
  \cos(z_R^b,z_T^{b_{R\rightarrow T}})
  \right),\\
\omega_b
&=\mathrm{clip}\!\left(
  \omega_b^{\mathrm{vis}}
  [1+\lambda_G(G_b-0.5)],0,1
  \right),\\
\mathcal{L}_{\mathrm{cma}}
&=
\frac{
  \sum_{b\in\mathcal{R}}\omega_b
  [1-\cos(z_R^b,z_T^{b_{R\rightarrow T}})]
}{
  \epsilon+\sum_{b\in\mathcal{R}}\omega_b
}.
\end{array}
\label{eq:cma}
\end{equation}
Here, $\sigma$ is the sigmoid function, $\mathrm{clip}$ bounds the weight to $[0,1]$, and $\epsilon$ ensures numerical stability.
The weight suppresses uncertain correspondence signals in the alignment objective rather than directly rescaling the fused features.
For neighboring source regions, let $\mathcal{N}=\{(i,j)\mid b_i,b_j\in\mathcal{R},0<\|p_i-p_j\|_2<r_s\}$, where $p_i$ is the center of $b_i$.
We define $\mathcal{L}_{\mathrm{smooth}}=\sum_{(i,j)\in\mathcal{N}}\|\Delta_i^{R\rightarrow T}-\Delta_j^{R\rightarrow T}\|_2^2/\max(|\mathcal{N}|,1)$ to discourage abrupt offset changes within radius $r_s$.
The complete CRA objective is
\begin{equation}
\mathcal{L}_{\mathrm{CRA}}
=
\lambda_a\mathcal{L}_{\mathrm{align}}
+\lambda_c\mathcal{L}_{\mathrm{cyc}}
+\lambda_m\mathcal{L}_{\mathrm{cma}}
+\lambda_s\mathcal{L}_{\mathrm{smooth}}.
\label{eq:cra}
\end{equation}
CRA consequently combines limited box-level anchors with cycle consistency and reliability-aware matching.
The regularized cross-modal representation provides the basis for the fused consensus in CMC-MT without adding an inference-time alignment branch.

\subsection{Cross-Modal Consensus Mean-Teacher}
CRA regularizes regional correspondence, but the few labeled pairs remain insufficient for learning an accurate detector.
Moreover, a missed target in either modality can produce incomplete supervision, and this error may circulate through the student--teacher update.
CMC-MT addresses this problem by generating pseudo labels from fused cross-modal evidence under pair-preserving views.

\noindent\textbf{Pair-Preserving Teacher--Student Views.}
For each unlabeled pair $(I_R,I_T)$, the EMA teacher model receives a weak view and the trainable student model receives a strong view~\cite{sohn2020fixmatch,liu2021unbiased,liu2022unbiasedv2}.
Both modalities share the same geometric transformation $\mathcal{H}$, while modality-specific photometric operators produce the weak and strong observations:
\begin{equation}
I_m^{v}=\mathcal{P}_m^{v}\!\left(\mathcal{H}(I_m)\right),
\qquad
m\in\{R,T\},\quad v\in\{w,s\}.
\label{eq:paired_views}
\end{equation}
Thus, the RGB--IR geometry is preserved within each view, and teacher boxes remain valid for the student view because the weak and strong observations use the same $\mathcal{H}$.
This paired construction avoids the coordinate inconsistency that would arise if either modality were transformed independently.

\noindent\textbf{Cross-Modal Consensus Supervision.}
The teacher processes the weak pair and aggregates the CRA-regularized RGB and infrared evidence through its fused branch.
Let $\Phi_{\bar{\theta}}$ and $D_{\bar{\theta}}$ denote the teacher fusion and oriented detection functions.
The consensus pseudo-label set is
\begin{equation}
\begin{array}{rl}
\mathcal{P}_{\mathrm{con}}^{w}
&=
D_{\bar{\theta}}\!\left(
\Phi_{\bar{\theta}}(I_R^{w},I_T^{w})
\right),\\
\widetilde{\mathcal{B}}_{\mathrm{con}}
&=
\left\{(b,c)\in\mathcal{P}_{\mathrm{con}}^{w}
\ \middle|\ s_b\geq\tau\right\}.
\end{array}
\label{eq:consensus_pseudo}
\end{equation}
RGB and infrared responses are complementary, so the fused prediction can retain target evidence when an individual branch misses or weakly observes an object.
The resulting consensus labels can consequently be more complete than supervision drawn from either displaced branch alone.
They supervise the student detector on the strong paired view after a short burn-in stage.
The teacher is updated as the exponential moving average of the student after each optimization step~\cite{tarvainen2017mean}.
This feedback loop turns completely unlabeled pairs into training targets, while fused consensus prevents branch-wise misses from recurring through self-training.

\subsection{Text-Guided Cross-Modal Instance Augmentation}
CMC-MT exploits unlabeled pairs, but it cannot create additional ground-truth instances for tail classes.
As the labeling budget decreases, tail-class annotations become critically scarce.
Conventional single-modal instance composition~\cite{ghiasi2021simple} can increase instance counts, but it does not preserve the correspondence of an RGB--IR pair.
TG-CMIA couples reliable tail-instance selection with scene-aware, offset-preserving pair composition.

\noindent\textbf{Consensus-Guided Tail-Instance Sampling.}
TG-CMIA first constructs an instance bank from the labeled pairs.
An RGB--IR crop pair $\iota_i$ is retained only when the two annotations agree in class and their centers satisfy the expected cross-modal displacement:
\begin{equation}
\mathcal{B}
=
\left\{
\iota_i\
\middle|\
\begin{array}{c}
c_R^i=c_T^i\in\mathcal{C}_{\mathrm{tail}},\\[-2pt]
\|p_R^i-p_T^i\|_2<\tau_d\max(w_i,h_i)
\end{array}
\right\}.
\label{eq:consensus_bank}
\end{equation}
Here, $\mathcal{C}_{\mathrm{tail}}=\{\mathrm{truck},\mathrm{freight\ car},\mathrm{van}\}$ denotes the three target categories used by TG-CMIA.
The variables $p_R^i$ and $p_T^i$ are the paired box centers, $(w_i,h_i)$ is the size of the RGB source box, and $\tau_d=0.6$.
Each bank entry stores the paired crops, their oriented boxes, and their natural relative offset.
Thus, consensus denotes agreement between the paired annotations rather than a teacher prediction, and it filters inconsistent instance pairs before augmentation.

To direct the limited composition budget toward scarce categories, an instance $\iota_i$ is sampled according to
\begin{equation}
P(\iota_i)
=
\frac{n_{c_i}^{-1}}
{\sum_{\iota_j\in\mathcal{B}}n_{c_j}^{-1}},
\label{eq:inverse_sampling}
\end{equation}
where $n_{c_i}$ is the number of bank instances from class $c_i$.
The inverse-frequency distribution increases the sampling probability of the rarest available classes without altering the original labels.
It follows the general principle of frequency-aware rebalancing~\cite{cui2019classbalanced}.

\begin{table*}[!t]
\begin{minipage}[t]{0.485\textwidth}
\centering
\captionof{table}{DroneVehicle results across labeling budgets.}
\label{tab:drone-main}
\begingroup
\setlength{\tabcolsep}{4.0pt}\small
\begin{tabular}{lcccc}
\toprule
Method & Input & 1\% & 5\% & 10\%\\
\midrule
\multicolumn{5}{l}{\emph{Supervised}}\\
C\textsuperscript{2}Former~\cite{yuan2024c2former} & V+I & 0.299 & 0.485 & 0.575\\
DMM~\cite{zhou2025dmm}        & V+I & 0.366 & \underline{0.599} & \underline{0.663}\\
SM3Det~\cite{li2026sm3det}     & I   & 0.389 & 0.548 & 0.609\\
M2D-LIF~\cite{zhao2025m2dlif}    & V+I & 0.271 & 0.491 & 0.602\\
\midrule
\multicolumn{5}{l}{\emph{Semi-supervised}}\\
MT~\cite{tarvainen2017mean} & V+I & 0.452 & 0.578 & 0.612\\
SOOD~\cite{hua2023sood}          & I & 0.476 & 0.581 & 0.594\\
DT~\cite{zhou2022denseteacher} & I & 0.477 & 0.594 & 0.621\\
MCL~\cite{wang2025mcl}           & I & \underline{0.492} & 0.592 & 0.614\\
\textbf{ACT (ours)} & V+I & \textbf{0.511} & \textbf{0.644} & \textbf{0.683}\\
\bottomrule
\end{tabular}
\endgroup
\end{minipage}\hfill
\begin{minipage}[t]{0.485\textwidth}
\centering
\captionof{table}{VEDAI results across labeling budgets.}
\label{tab:vedai-main}
\begingroup
\setlength{\tabcolsep}{4.0pt}\small
\begin{tabular}{lcccc}
\toprule
Method & Input & 5\% & 10\% & 15\%\\
\midrule
\multicolumn{5}{l}{\emph{Supervised}}\\
C\textsuperscript{2}Former~\cite{yuan2024c2former} & V+I & 0.130 & 0.286 & 0.342\\
DMM~\cite{zhou2025dmm}        & V+I & 0.090 & 0.221 & 0.272\\
SM3Det~\cite{li2026sm3det}     & I   & 0.127 & 0.329 & 0.484\\
M2D-LIF~\cite{zhao2025m2dlif}    & V+I & 0.064 & 0.080 & 0.158\\
\midrule
\multicolumn{5}{l}{\emph{Semi-supervised}}\\
MT~\cite{tarvainen2017mean} & V+I & 0.371 & \underline{0.444} & 0.488\\
SOOD~\cite{hua2023sood}          & I & 0.335 & 0.428 & \underline{0.506}\\
DT~\cite{zhou2022denseteacher} & I & 0.284 & 0.440 & 0.366\\
MCL~\cite{wang2025mcl}           & I & \underline{0.415} & 0.443 & 0.483\\
\textbf{ACT (ours)} & V+I & \textbf{0.541} & \textbf{0.545} & \textbf{0.632}\\
\bottomrule
\end{tabular}
\endgroup
\end{minipage}
\end{table*}

\noindent\textbf{Scene-Aware Offset-Preserving Composition.}
Vision-language pretraining provides semantic scene representations without task-specific box labels~\cite{radford2021clip}.
For each target pair, a frozen Qwen2.5-VL-32B model~\cite{bai2025qwen25vl} generates the scene prior $\mathcal{S}=(G,d,\rho)$ once without annotation input.
Here, $G$ is a $2\times2$ occupancy grid, $d$ is a day--night descriptor, and $\rho$ contains the two modality reliabilities.
During detector training, $G$ selects the composition cell, $d$ permits one additional inserted pair at night, and $\rho$ scales the modality-specific feathering masks.
Once an instance pair is selected, TG-CMIA applies the same translation $t$ to its RGB and infrared box centers:
\begin{equation}
\begin{array}{c}
p_R'=p_R+t,\qquad p_T'=p_T+t,\\
p_T'-p_R'=p_T-p_R.
\end{array}
\label{eq:offset_preserving}
\end{equation}
The second relation shows that the original cross-modal displacement is preserved after composition.
The paired crops are blended into their corresponding regions without changing their relative scale or orientation, and the transformed oriented boxes are appended to both annotation sets.
A composed instance may remain faint in nighttime RGB imagery while appearing clear in infrared, yet the two annotations stay geometrically paired.
TG-CMIA thus adds targeted ground-truth supervision for tail classes without introducing cross-modal inconsistency.

\noindent\textbf{Overall Training Objective.}
Let $\mathcal{D}_l$ denote the labeled pair set and let
$\mathcal{D}_l^{+}=\mathcal{A}_{\mathrm{TG}}(\mathcal{D}_l;\mathcal{B},\mathcal{S})$
denote its TG-CMIA-augmented counterpart.
The complete student objective is
\begin{equation}
\mathcal{L}_{\mathrm{ACT}}
=
\mathcal{L}_{\mathrm{det}}^{l}(\mathcal{D}_l^{+})
+\lambda_u\mathcal{L}_{\mathrm{det}}^{u}
  (\widetilde{\mathcal{B}}_{\mathrm{con}})
+\mathcal{L}_{\mathrm{CRA}},
\label{eq:act_objective}
\end{equation}
where the three terms correspond to supervised detection on the augmented labeled pairs, consensus supervision on the unlabeled pairs, and cross-modal alignment, respectively.
The student is optimized by Eq.~(\ref{eq:act_objective}), while the teacher is updated as the exponential moving average of the student.

\section{Experiments}

\subsection{Datasets and Evaluation Protocol}
\textbf{DroneVehicle.} DroneVehicle~\cite{sun2022dronevehicle} contains 28{,}439 RGB--IR image pairs collected from daytime and nighttime scenes. It covers five vehicle categories with oriented annotations in both modalities. The official split contains 17{,}990 training, 1{,}469 validation, and 8{,}980 test pairs.

\noindent\textbf{VEDAI.} VEDAI~\cite{razakarivony2016vedai} covers four vehicle categories. Our split contains 772 training pairs and 192 validation pairs. Its average object width is approximately 14 pixels, compared with 50 pixels on DroneVehicle. We use it to evaluate ACT under limited data and substantially smaller objects.

\noindent\textbf{Image-pair-level protocol.} For each budget $p$, we select a fixed subset containing $p\%$ of the training image pairs. Both modalities are labeled in the selected pairs, while all annotations are removed from the remaining pairs. We use $p\in\{1,5,10\}$ on DroneVehicle and $p\in\{5,10,15\}$ on VEDAI. These ratios correspond to 180, 900, and 1{,}799 labeled pairs on DroneVehicle, and 38, 77, and 115 labeled pairs on VEDAI.

\noindent\textbf{Evaluation metric.} Following semi-supervised detection practice~\cite{hua2023sood,xu2021softteacher}, we evaluate against infrared validation annotations and report rotated mAP at $0.5$ IoU with 11-point AP interpolation~\cite{everingham2010pascal}. All methods use identical splits and evaluation code.

\subsection{Implementation Details}
\textbf{Detector architecture.} The detector uses two ResNet-50 streams~\cite{he2016resnet}. At each backbone stage, adaptive feature sampling and inter-modality cross-attention calibrate and exchange complementary features~\cite{yuan2024c2former}. The resulting multi-scale representations are fused by an FPN~\cite{lin2017fpn} and passed to an S\textsuperscript{2}A-Net oriented detection head~\cite{han2021s2anet}.

\noindent\textbf{Training settings.} We implement ACT with MMRotate 0.3.4~\cite{zhou2022mmrotate} on one NVIDIA A800 GPU. SGD uses a $0.001$ initial learning rate, $0.9$ momentum, $0.0001$ weight decay, and 500-iteration linear warm-up. Each iteration samples one labeled and one unlabeled pair at $512\times640$ on DroneVehicle or $1024\times1024$ on VEDAI. DroneVehicle is trained for six epochs with decays at epochs 4 and 5. VEDAI uses twelve epochs with decays at epochs 8 and 11. Both add two refinement epochs at learning rate $0.0005$.

\noindent\textbf{Semi-supervised configuration.} Labeled and unlabeled pairs share resizing and geometric flips. Pair-preserving weak and strong views differ only in photometric augmentation. The teacher follows the student via EMA with $0.999$ momentum. After a $2{,}000$-iteration burn-in, fused predictions above $0.4$ confidence supervise strong views with weight $\lambda_u=1.0$. We evaluate thresholds in $\{0.3,0.4,0.5,0.7\}$ and select $0.4$; complete run details are supplementary. The CRA weights $(\lambda_a,\lambda_c,\lambda_m,\lambda_s)$ are $(1.0,0.5,0.5,0.3)$, and $\lambda_G=0.3$. Sparse anchors require at least $0.5$ cross-modal IoU, and the neighborhood radius $r_s$ is 64 pixels. With probability $0.8$, TG-CMIA adds one to four target-class pairs and at most one extra pair at night.

\subsection{Comparison with State-of-the-Art}
\begin{figure*}[!t]\centering
\includegraphics[width=\textwidth]{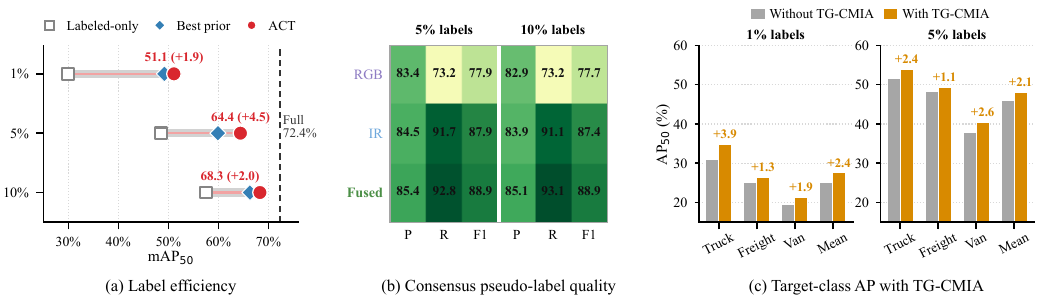}
\caption{Quantitative analysis on DroneVehicle. (a) Label efficiency across annotation ratios. (b) Teacher pseudo-label quality across RGB, IR, and fused branches. (c) Class-wise effects of TG-CMIA under the two lowest labeling budgets.}
\label{fig:quantitative-analysis}
\end{figure*}

\noindent\textbf{Comparison protocol.} Tables~\ref{tab:drone-main} and~\ref{tab:vedai-main} compare supervised detectors trained only on labeled subsets with semi-supervised detectors that also use unlabeled pairs. All methods share the same subsets and evaluation code. MT and DT denote Mean-Teacher and Dense Teacher; I and V+I denote infrared-only and paired RGB--IR input, respectively. Bold and underlined values mark the best and second-best results. For resolution-sensitive DMM, we report native-resolution VEDAI scores.

\noindent\textbf{Results on DroneVehicle.} ACT obtains $0.511$, $0.644$, and $0.683$ mAP at the $1\%$, $5\%$, and $10\%$ budgets. It outperforms the strongest semi-supervised competitor by $0.019$, $0.050$, and $0.062$, respectively, and surpasses every supervised detector, including DMM, SM3Det, and M2D-LIF. Its gain over C\textsuperscript{2}Former widens from $+0.108$ at $10\%$ to $+0.212$ at $1\%$, demonstrating the growing value of unlabeled pairs as annotations shrink. With $10\%$ labeled pairs, ACT recovers $94.3\%$ of the $0.724$ full-supervision mAP.

\noindent\textbf{Results on VEDAI.} ACT obtains $0.541$, $0.545$, and $0.632$ mAP, exceeding the corresponding second-best results by $0.126$, $0.101$, and $0.126$. With only 38 labeled pairs at the $5\%$ budget, it already surpasses every competing $15\%$ result, whose best mAP is $0.506$. Supervised detectors degrade sharply with only 38--115 labeled pairs, whereas ACT exploits all remaining unlabeled pairs. This advantage generalizes to smaller objects and varied alignment.

\noindent\textbf{Label efficiency.} Figure~\ref{fig:quantitative-analysis}(a) shows the largest gain at $1\%$; at $10\%$, ACT attains $94.3\%$ of full supervision.

\begin{table}[t]\centering
\caption{Per-category AP on DroneVehicle at the $10\%$ annotation ratio.}
\label{tab:perclass}
\begingroup
\setlength{\tabcolsep}{1.3pt}\fontsize{8.0}{9.2}\selectfont
\begin{tabular}{@{}lcccccc@{}}
\toprule
Method & car & truck & bus & van & freight & mAP\\
\midrule
C\textsuperscript{2}Former~\cite{yuan2024c2former} & 0.896 & 0.442 & 0.872 & 0.340 & 0.326 & 0.575\\
DMM~\cite{zhou2025dmm}        & \textbf{0.900} & \underline{0.555} & \underline{0.888} & \underline{0.449} & \underline{0.521} & \underline{0.663}\\
SM3Det~\cite{li2026sm3det}     & 0.897 & 0.474 & 0.881 & 0.360 & 0.434 & 0.609\\
M2D-LIF~\cite{zhao2025m2dlif}    & 0.889 & 0.508 & 0.778 & 0.412 & 0.423 & 0.602\\
MT~\cite{tarvainen2017mean} & 0.890 & 0.492 & 0.880 & 0.364 & 0.435 & 0.612\\
SOOD~\cite{hua2023sood}       & 0.897 & 0.462 & 0.862 & 0.367 & 0.383 & 0.594\\
DT~\cite{zhou2022denseteacher} & 0.897 & 0.492 & 0.868 & 0.413 & 0.441 & 0.622\\
MCL~\cite{wang2025mcl}        & \underline{0.898} & 0.467 & 0.864 & 0.399 & 0.442 & 0.614\\
\textbf{ACT (ours)} & 0.896 & \textbf{0.612} & \textbf{0.896} & \textbf{0.482} & \textbf{0.528} & \textbf{0.683}\\
\bottomrule
\end{tabular}
\endgroup
\end{table}

\noindent\textbf{Per-category comparison.} Table~\ref{tab:perclass} reports AP at the $10\%$ budget. Car and bus are near saturation, while the main differences occur on truck, van, and freight car. ACT ranks first on all three target categories and outperforms DMM by $0.057$, $0.033$, and $0.007$, respectively. Infrared-only semi-supervised methods lag further behind, supporting the use of paired cues and targeted TG-CMIA supervision.

\subsection{Ablation and Component Analysis}
\begin{table}[t]\centering
\caption{Component ablation on DroneVehicle.}
\label{tab:ablation}
\begingroup
\setlength{\tabcolsep}{2.5pt}\fontsize{8.0}{9.2}\selectfont
\begin{tabular*}{\columnwidth}{@{\extracolsep{\fill}}ccccccc@{}}
\toprule
\multicolumn{3}{c}{Components} & \multicolumn{4}{c}{Annotation ratio}\\
\cmidrule(lr){1-3}\cmidrule(lr){4-7}
CMC-MT & TG-CMIA & CRA & 1\% & 5\% & 10\% & 100\%\\
\midrule
 & & & 0.299 & 0.485 & 0.575 & \emph{0.724}\\
\cmark & & & 0.489 & 0.619 & 0.665 & --\\
\cmark & \cmark & & 0.510 & 0.639 & 0.679 & --\\
\cmark & \cmark & \cmark & \textbf{0.511} & \textbf{0.644} & \textbf{0.683} & --\\
\bottomrule
\end{tabular*}
\endgroup
\end{table}

\noindent\textbf{Overall component contribution.} Table~\ref{tab:ablation} sequentially adds the three mechanisms to the supervised baseline at the $1\%$, $5\%$, and $10\%$ budgets. Each row reports the peak result of that configuration; the $100\%$ column is only the full-supervision upper bound. ACT improves the limited-label baseline by $21.2$, $15.9$, and $10.8$ mAP points and reaches $70.6\%$, $89.0\%$, and $94.3\%$ of the reference, respectively. Because the mechanisms are added sequentially, each difference is a conditional rather than isolated causal gain.

\noindent\textbf{Component-wise analysis.}
\textit{Effect of CMC-MT.} CMC-MT brings the largest incremental gains of $19.0$, $13.4$, and $9.0$ mAP points. The gain is largest at the $1\%$ budget, where nearly all training pairs are unlabeled and reliable pseudo labels are therefore most valuable. Figure~\ref{fig:quantitative-analysis}(b) explains this behavior. Relative to the IR branch, fused consensus improves recall by $1.1$ and $2.0$ points and F1 by $1.0$ and $1.5$ points at the $5\%$ and $10\%$ budgets. Precision also increases by $0.9$ and $1.2$ points, so the recovered branch-wise misses do not come from indiscriminate box retention. Meanwhile, RGB recall trails fused consensus by $19.6$ and $19.9$ points. These diagnostics connect complementary cross-modal recovery to more reliable supervision on unlabeled pairs.

\textit{Effect of TG-CMIA.} TG-CMIA adds $2.1$, $2.0$, and $1.4$ mAP points. Figure~\ref{fig:quantitative-analysis}(c) shows that the target-class mean rises from $0.252$ to $0.276$ at $1\%$ and from $0.459$ to $0.480$ at $5\%$. Its larger gain at $1\%$ is consistent with more severe tail-class scarcity under the lowest budget. At $10\%$, it rises from $0.511$ to $0.522$, while car and bus remain nearly unchanged. On the nighttime subset, TG-CMIA adds $1.4$ mAP points and improves van AP by $3.7$ points.

\textit{Effect of CRA.} Finally, CRA adds $0.1$, $0.5$, and $0.4$ mAP points across the three budgets. Because it is added after CMC-MT and TG-CMIA, these values measure its conditional contribution to the complete pipeline. The smaller gain at $1\%$ is consistent with the scarcity of labeled anchors at the lowest budget. Nevertheless, its improvement at every budget shows that regional correspondence remains complementary to consensus supervision. CRA regularizes the fused representation from which CMC-MT derives pseudo labels, linking better regional correspondence to the subsequent self-training stage.

\begin{figure}[!t]\centering
\includegraphics[width=\columnwidth]{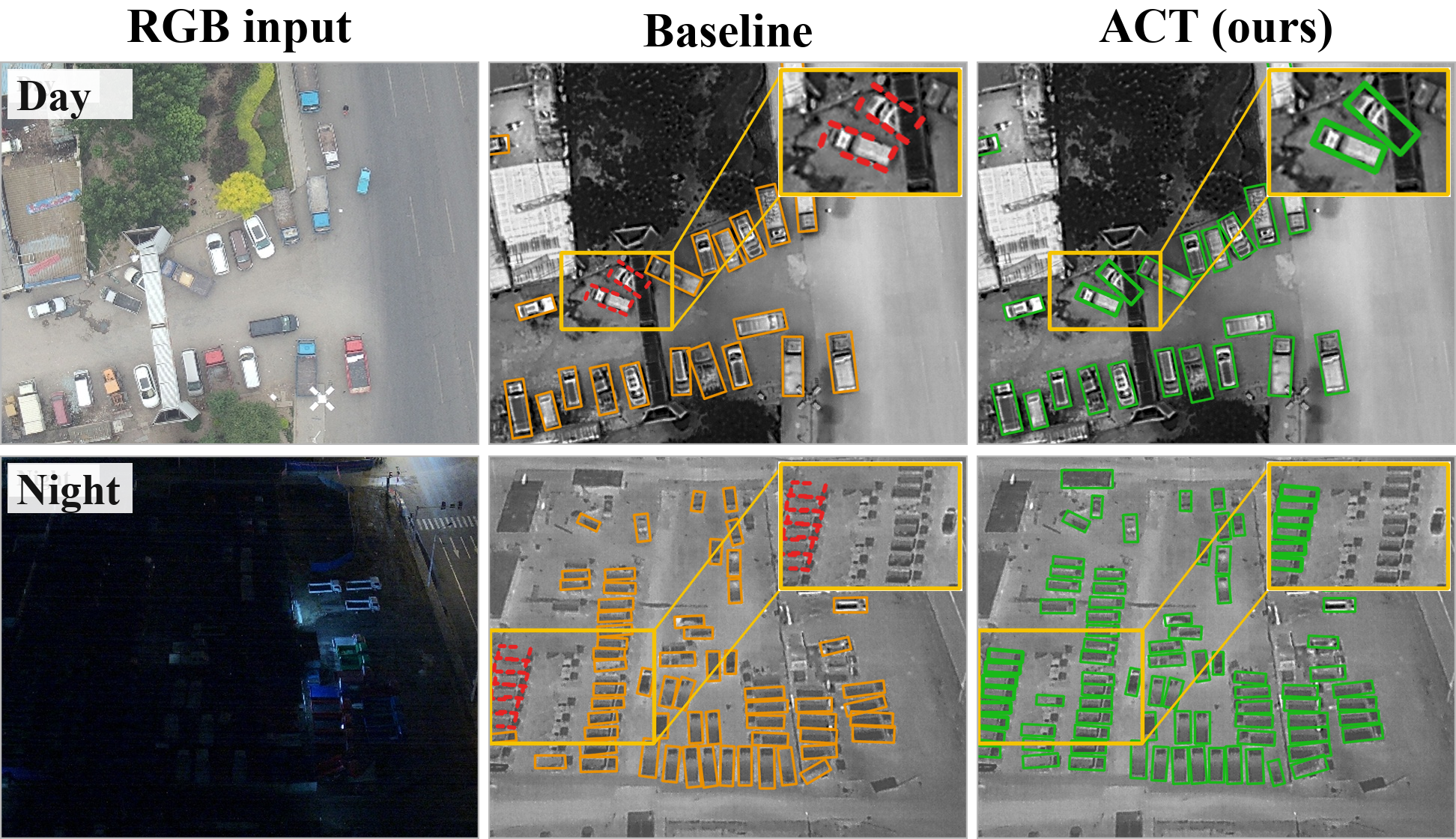}
\caption{Qualitative results on DroneVehicle. Rows show day and night scenes; columns show RGB input, baseline predictions (orange), and ACT predictions (green). Red dashed boxes mark targets recovered by ACT.}
\label{fig:qual}
\end{figure}

\subsection{Robustness and Efficiency}
\noindent\textbf{Robustness across seeds.} The main comparison follows the same single-run peak protocol for all methods. As a robustness check, three runs at the representative $10\%$ budget obtain $0.681\pm0.001$ mAP, only $0.002$ below the headline result. Complete per-seed results at the $1\%$, $5\%$, and $10\%$ budgets are provided in the supplementary material.

\noindent\textbf{Pseudo-label threshold sensitivity.}
In a controlled same-seed six-epoch study at the $10\%$ budget, $\tau=0.3$, $0.4$, $0.5$, and $0.7$ yield $0.647$, $0.674$, $0.657$, and $0.658$ mAP, respectively.
All four runs use the same labeled subset and training schedule, with only $\tau$ changed.
Relative to the selected value, the other settings reduce mAP by $2.7$, $1.7$, and $1.6$ points, respectively.
Thus, $\tau=0.4$ provides the best empirical balance between pseudo-label coverage and reliability.
We retain this value across annotation budgets rather than tuning it separately for each reported result.

\noindent\textbf{Reference-modality consistency.}
Primary results use infrared annotations.
To rule out dependence on this reference, we also evaluate the same $10\%$ models against RGB annotations.
ACT obtains $0.645$ mAP versus $0.528$ for its supervised baseline, preserving an $11.7$-point advantage despite spatial RGB--IR displacement.
Thus, the conclusion remains consistent across both reference modalities.

\noindent\textbf{Efficiency.} ACT optimizes $150.19$M parameters: $120.83$M belong to the deployed fused detector, while $29.36$M come from training-only alignment and auxiliary modality paths. TG-CMIA also operates only during training and adds no inference parameters. At inference, ACT retains the EMA teacher and its fused branch, matching the supervised baseline's graph. Both models therefore use $120.83$M parameters and reach $3.2$ FPS (batch size one, $512\times640$) on an A800 GPU. Thus, the additional mechanisms increase training cost without enlarging the deployment footprint.

\subsection{Qualitative Detection Analysis}
Figure~\ref{fig:qual} compares ACT with the supervised baseline in daytime and nighttime scenes.
In the daytime example, the baseline misses several tightly packed vehicles, whereas ACT recovers more of the group with orientations consistent with the road layout.
At night, the RGB observation is nearly dark, but the corresponding IR view retains clear vehicle responses; ACT consequently produces more complete detections in the parking area.
The highlighted recoveries show that complementary cross-modal evidence benefits both crowded layouts and severe illumination degradation.
They also localize the improvement to genuine baseline misses rather than merely denser predictions on already detected objects.
The two scenes therefore separate recovery in a crowded daytime layout from recovery under severe nighttime appearance degradation, complementing the aggregate accuracy reported in the tables.

\noindent\textbf{Limitations.}
ACT assumes that RGB and IR frames are available as synchronized pairs.
CRA accommodates object-level displacement, but it does not remove this pairing requirement.
Moreover, the current evaluation focuses on aerial vehicle categories, and tail classes remain difficult when nighttime RGB evidence is severely degraded.
Performance under missing modalities, stronger temporal misalignment, and broader object taxonomies has not yet been established.
These boundaries motivate extensions to incomplete sensor inputs and stronger vision-language priors.

\section{Conclusion}
This paper studies image-pair-level semi-supervised VIOD, where few RGB--IR pairs are jointly labeled and the rest are unlabeled. ACT combines CRA, CMC-MT, and TG-CMIA to address insufficient alignment supervision, accumulated pseudo-label noise, and tail-class scarcity, respectively. Experiments on DroneVehicle and VEDAI show consistent gains, especially at low labeling budgets. The strongest gains confirm the value of unlabeled pairs when paired supervision is scarce. With 10\% labeled pairs on DroneVehicle, ACT retains 94.3\% of full-supervision mAP, demonstrating label efficiency across datasets and scene conditions.

Regional correspondence and fused consensus enable label-efficient self-training, while paired augmentation alleviates tail-class scarcity. Nighttime tail classes remain challenging under severely degraded RGB observations, motivating stronger remote-sensing vision-language priors.

\begingroup
\small
\sloppy
\bibliographystyle{ieee_fullname}
\bibliography{act}
\endgroup

\end{document}